%% file: main.tex
\documentclass[10pt,twocolumn,letterpaper]{article}
\usepackage[pagenumbers]{cvpr}
\input{preamble}

\definecolor{cvprblue}{rgb}{0.21,0.49,0.74}
\usepackage[breaklinks,colorlinks,allcolors=cvprblue]{hyperref}
\input{paper_info}
\hypersetup{pdftitle={Few-Shot Adaptation of Vision-Language Models via Agentic Model Editing},
  pdfauthor={Yusheng Zhao, Xiao Luo, Zhiping Xiao, Ming Zhang}}
\begin{document}
\begingroup
\setlength{\tabcolsep}{4pt}
\maketitle
\endgroup
\input{0_abstract}
\input{1_intro}
\input{2_related}

\input{3_method}
\input{4_experiments}

\input{5_conclusion}
\FloatBarrier
{\small
\bibliographystyle{ieeenat_fullname}
\bibliography{ref}
}
\end{document}

%% file: preamble.tex
\usepackage[T1]{fontenc}
\usepackage{mathtools}
\usepackage{bm}
\usepackage{pifont}
\usepackage{multirow}
\usepackage{tabularx}
\usepackage{makecell}
\usepackage{placeins}
\usepackage{needspace}
\usepackage[ruled,vlined]{algorithm2e}

\input{math_commands}
\graphicspath{{figures/}}

\usepackage{colortbl}
\input{tables/main_table_style}

%% file: math_commands.tex
\DeclareMathOperator*{\argmax}{arg\,max}

%% file: tables/main_table_style.tex
\definecolor{ameTableHead}{RGB}{198,223,181}
\definecolor{ameTableStripe}{RGB}{242,242,242}
\definecolor{ameTableOurs}{RGB}{252,242,196}
\newcommand{\ameScore}[2]{#1\textsubscript{\fontsize{6}{7}\selectfont\textpm#2}}

%% file: paper_info.tex
\def\confName{CVPR}
\def\confYear{2027}
\newcommand{\method}{{%
  \fontfamily{lmtt}\fontseries{b}%
  \ifdim\fontdimen1\font>0pt\relax
    \fontshape{sl}%
  \else
    \fontshape{n}%
  \fi
  \selectfont AME}\xspace}
\newcommand{\modelname}[1]{{\fontfamily{pcr}\selectfont #1}\xspace}
\newcommand{\papertitle}{Few-Shot Adaptation of Vision-Language Models via Agentic Model Editing}
\title{\papertitle}
\author{%
Yusheng Zhao\\
{\small Peking University}\\
{\ttfamily\footnotesize yushengzhao99@gmail.com}
\and
Xiao Luo\\
{\small University of Wisconsin--Madison}\\
{\ttfamily\footnotesize xiao.luo@wisc.edu}
\and
Zhiping Xiao\\
{\small University of Washington}\\
{\ttfamily\footnotesize patxiao@uw.edu}
\and
Ming Zhang\\
{\small Peking University}\\
{\ttfamily\footnotesize mzhang\_cs@pku.edu.cn}
}

%% file: 0_abstract.tex
\begin{abstract}
Few-shot adaptation transfers pretrained Vision-Language Models (VLMs) to new domains using a small set of labeled images. Most methods fit prompts, representations, or classifiers within an adaptation procedure specified in advance. Yet the visual distinctions needed for recognition can differ substantially across domains, motivating the question: \textbf{\textit{Can an LLM agent use few-shot examples and model observations to design the adaptation procedure itself?}} We propose \underline{A}gentic \underline{M}odel \underline{E}diting (\method), an agentic framework that jointly designs input transformations, encoder edits, feature computation, and classifier fitting as an executable, domain-customized adaptation program. The agent inspects images, class semantics, model parameters, intermediate representations, and activations to formulate hypotheses about the target task. It implements and tests candidate programs on the support set, with aggregate validation accuracy guiding refinement and selection. Each program fits a complete domain model that is fixed before testing. The agent and its tools are shared across domains, while each domain receives its own adaptation program. Across eleven datasets with \modelname{CLIP ViT-B/16}, \method achieves the highest average accuracies of $81.2\%$ without back-propagation and $81.9\%$ with back-propagation among the compared methods. Further experiments demonstrate applicability across VLM backbones and LLM agents.
\end{abstract}

%% file: 1_intro.tex
\section{Introduction}
\label{sec:intro}

Vision-Language Models (VLMs) \cite{radford2021learning, dai2023instructblip, bordes2024introduction, zhang2024vision} connect visual recognition with natural language, enabling image classification through textual prompts.
Their pretrained representations, however, may not capture the distinctions needed in a target domain whose image distribution differs from pretraining \cite{sung2022vl, wu2024transferring, gao2024CLIPAdapter, zhao2025ABPEM}.
Few-shot adaptation \cite{ouali2023black, silvarodriguez2024CLAP, farina20252SFS} addresses this mismatch using a small set of labeled target examples. Existing approaches adapt prompts \cite{zhou2022CoOp,khattak2023MaPLe, zheng2025HiCroPL}, representations \cite{gao2024CLIPAdapter, farina20252SFS}, classifiers \cite{silvarodriguez2024CLAP, li2025TIMO}, or feature matching \cite{zhu2023APE, ye2026CDPS} to improve recognition from the few labeled examples available in each target domain.

\begin{figure}[t]
    \centering
    \includegraphics[width=\linewidth]{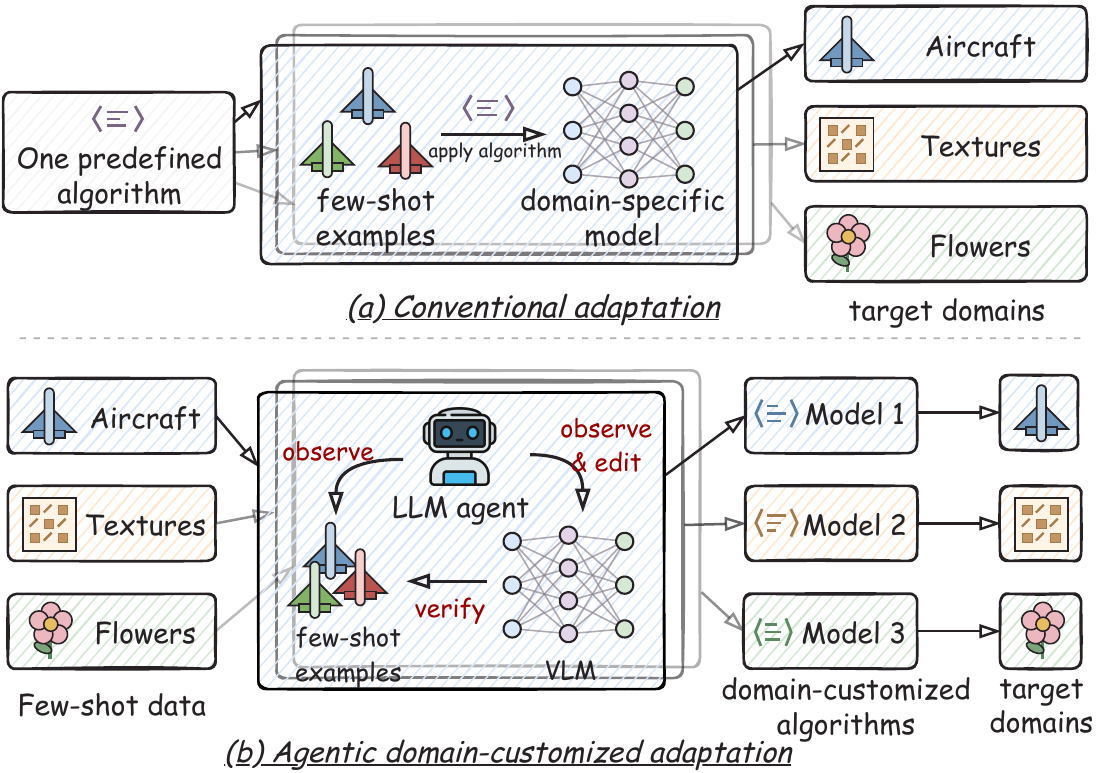}
    \caption{Conventional few-shot adaptation applies a predefined algorithm across domains (a). \method instead uses a shared LLM agent to design a domain-customized adaptation program from labeled examples and observations of the pretrained model (b).}
    \vspace{-3mm}
    \label{fig:motivation}
\end{figure}

Despite their progress, conventional methods \cite{zhang2022TipAdapter, udandarao2023TIPX,wang2024GDA, fahes2026ProLIP} generally apply one predefined adaptation procedure across domains (Figure~\ref{fig:motivation}a). Target examples determine the fitted parameters or prompts, but the rules for feature computation, parameter updates, and classifier construction are specified in advance.
Different domains can require different recognition strategies \cite{konishi2000statistical, park2011domain, bui2021exploiting}. Distinguishing aircraft variants may depend on subtle structural details, whereas recognizing textures may rely on repeated local patterns and their statistics. This raises a design question beyond parameter fitting: which operations should an adaptation algorithm use for a particular domain?

We therefore study \emph{domain-customized adaptation programs}. Few-shot examples provide both supervision and evidence for designing these programs: images reveal visual characteristics, and class names identify the distinctions the task requires.
Model observations, including intermediate representations, activations, and prediction errors, reveal how those distinctions are represented or lost during computation. Together, these observations can guide feature selection, identify encoder components to edit, and inform classifier construction for the target task.

Large Language Models (LLMs) \cite{grattafiori2024llama, olmo2025olmo, anthropic2026ClaudeOpus55, zhao2026SwitchAttention, zhao2025DENSE} offer capabilities relevant to this design problem. Multimodal perception \cite{zhan2024anygpt} supports inspection of target examples, instruction understanding \cite{qin2024infobench} connects observations to the task, and code generation \cite{dong2025survey, tang2026autoagent} turns hypotheses into executable procedures. We ask:
\begin{quotation}
\textbf{\textit{How can an LLM agent use few-shot examples and model observations to design a domain-customized adaptation algorithm?}}
\end{quotation}

We propose \underline{A}gentic \underline{M}odel \underline{E}diting (\method), an agentic framework for designing executable few-shot adaptation programs.
As illustrated in Figure~\ref{fig:motivation}b, the agent uses labeled images, class semantics, and internal VLM observations to propose changes in how target images are represented and classified. It translates these hypotheses into an executable fitting procedure that specifies input transformations, encoder weight updates, feature computation, and classifier construction together. Support-set experiments and aggregate validation feedback guide successive revisions. The selected procedure produces a fixed predictor for the target domain, requiring no agent interaction at inference.
The agent, instructions, and tools are shared across domains. Within each domain, encoder edits determine the representations available to subsequent statistics and classifiers, whose predictions guide further edits. The agent designs these operations together as one adaptation program with limited manual involvement.
Our contributions are threefold:
\begin{enumerate}
    \item[\ding{182}] \textit{\textbf{New Perspective.}} We study few-shot VLM adaptation as domain-conditioned program design. Target examples and model observations inform the adaptation procedure itself, including feature computation, parameter updates, and classifier construction, beyond fitting parameters within a predefined procedure.
    \item[\ding{183}] \textit{\textbf{Novel Methodology.}} We propose \method, which jointly designs encoder adaptation and classifier construction from multimodal and internal model observations. The agent develops executable adaptation programs through observation, model editing, and support-set experiments, guided by aggregate external feedback.
    \item[\ding{184}] \textit{\textbf{Extensive Experiments.}} We evaluate \method across eleven datasets against competing baselines, showing that \method is \textbf{(1) versatile}, operating both with and without back-propagation (BP), \textbf{(2) high performing}, achieving the highest average accuracies of 81.2\% (BP-free) and 81.9\% (BP-based), and \textbf{(3) broadly applicable}, adapting both contrastive and generative VLMs. 
\end{enumerate}

%% file: 2_related.tex
\begin{figure*}[t]
    \centering
    \includegraphics[width=\textwidth]{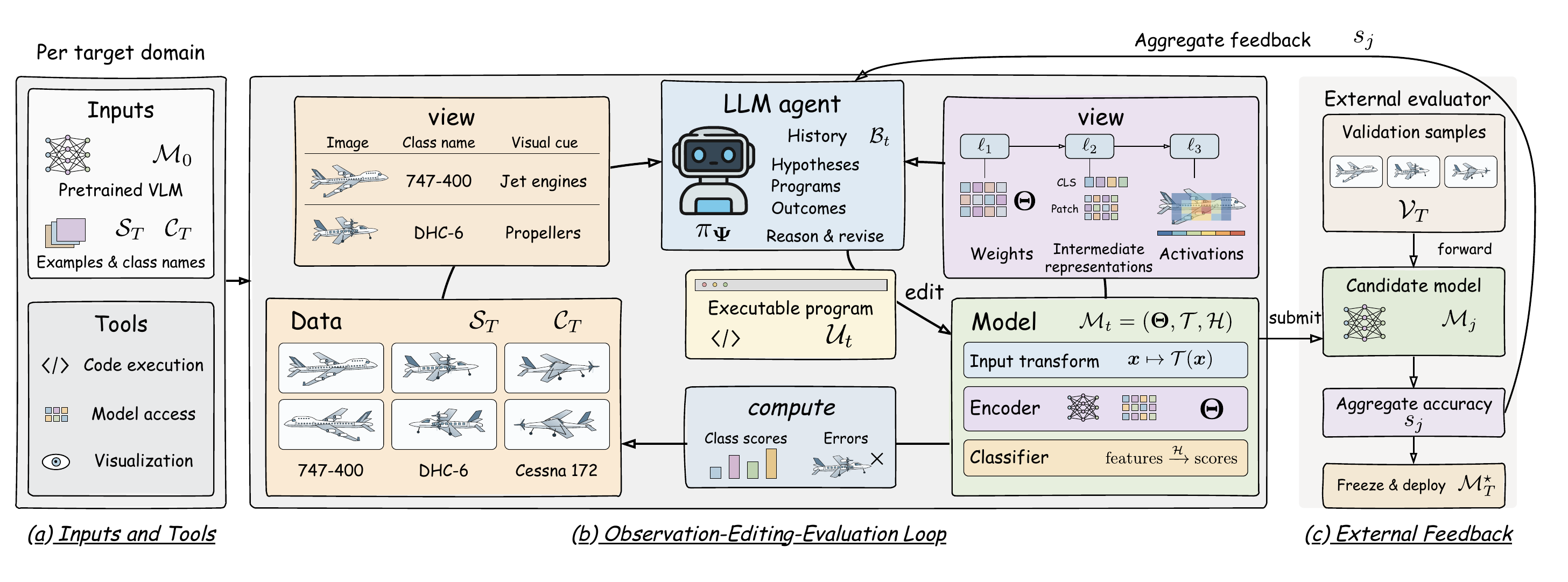}
    \vspace{-6mm}
    \caption{Overview of \method. (a) Inputs and tools expose labeled target examples and the pretrained VLM. (b) The agent inspects images, class semantics, weights, representations, and activations to formulate, implement, and test model-editing hypotheses on the support set. Each experiment guides the next round of observation and editing. (c) External feedback guides refinement and selection for deployment.}
    \vspace{-2mm}
    \label{fig:framework}
\end{figure*}
\section{Related Work}
\label{sec:related}
\subsection{Vision-Language Models}
Vision-Language Models (VLMs) \cite{radford2021learning,jia2021ALIGN,zhai2023SigLIP} often use image-text pairs to align visual and textual representations in a shared embedding space.
This alignment enables zero-shot classification by comparing an image embedding with text embeddings describing the candidate categories, without training a separate classifier for each downstream task \cite{radford2021learning,zhang2024vision}.
The language interface also allows additional semantic knowledge to be incorporated into recognition \cite{pratt2023CuPL}. 
However, the representations learned from large-scale pretraining data may not fully capture the visual distinctions required by a target domain \cite{wortsman2022WiSEFT}, and classification can be sensitive to the choice of textual descriptions \cite{zhou2022CoOp}.
Therefore, this work aims to adapt pretrained VLMs with a limited number of labeled samples from the target distribution.

\subsection{Few-Shot Adaptation of VLMs}
Few-shot adaptation \cite{motiian2017few, ye2020few, gao2026optimization, xiao2026CONST} transfers a pretrained model to a downstream task using limited labeled data.
\emph{Prompt-based methods} \cite{yao2023KgCoOp,zhu2023ProGrad,zheng2025HiCroPL} optimize the inputs to the pretrained encoders, often via learning continuous context vectors \cite{zhou2022CoOp}, generating input-conditional prompts \cite{zhou2022CoCoOp}, or jointly optimizing prompts for the visual and language branches \cite{khattak2023MaPLe}.
\emph{Representation adaptation} \cite{gao2024CLIPAdapter,fahes2026ProLIP,farina20252SFS} modifies how features are computed or transformed by adding customized feature adapters \cite{gao2024CLIPAdapter} or tuning the model's parameters \cite{farina20252SFS,zanella2024CLIPLoRA}.
\emph{Classifier adaptation} \cite{yu2023TaskRes,wang2024GDA,li2025TIMO} constructs or refines the decision function from labeled target examples. 
\emph{Feature matching} \cite{zhu2023APE,udandarao2023TIPX,ye2026CDPS} classifies queries through their relationships with cached support features.
While these methods adapt prompts \cite{zhou2022CoOp,zhou2022CoCoOp}, representations \cite{gao2024CLIPAdapter,fahes2026ProLIP}, or classifiers \cite{yu2023TaskRes,wang2024GDA} to the target data, their adaptation procedures remain predefined.
\method instead uses an LLM agent to generate adaptation programs from target examples and internal model observations.

\subsection{LLM Agents}
LLM agents combine reasoning, tool use, and feedback to solve tasks through interaction \cite{doshi2026towards, liu2026toolscope}.
Instruction following supports interpretation of task requirements \cite{qin2024infobench}; reasoning and multi-turn interaction support the accumulation of observations and revision of hypotheses \cite{yao2023ReAct,xiang2025guardagent,guo2026se}.
Code generation and execution turn hypotheses into testable procedures \cite{wang2024CodeAct,liu2026lessons, kang2026distilling}, while multimodal perception supports inspection of visual examples and their semantics \cite{zhan2024anygpt,yang2026mla}.
Recent work also explores LLM reasoning about neural computation and model parameters \cite{takehi2026weightspace}. These capabilities motivate an agent that investigates both target data and the pretrained model when designing an adaptation program.
Closely related approaches \cite{liu2024LBBO,du2024IPO} use LLMs to generate and evaluate domain-specific textual prompts, guided by performance feedback and, in some cases, visual descriptions.
Their search operates through the text interface of a fixed VLM. \method expands the executable adaptation space to input transformations, encoder weight updates, feature computation, and classifier construction. The agent can connect observations of target images and class semantics with internal representations and activations when deciding which operations to implement.

%% file: 3_method.tex
\section{Preliminaries}
Given a pretrained VLM $\mathcal M_0$, we consider adaptation to a target distribution $p_T(\bm{x},y)$ with label set $\mathcal Y_T$ and corresponding class names $\mathcal C_T$.
A small collection of labeled target samples, called the support set $\mathcal S_T=\{(\bm{x}_i,y_i)\}_{i=1}^{n_T}$, contains $k$ images per class, where $n_T=k|\mathcal Y_T|$.
The goal is to construct a domain-adapted model for classifying previously unseen target images, without access to the VLM's pretraining data.
We represent an adapted model as $\mathcal M=(\bm{\Theta},\mathcal T,\mathcal H)$, where $\bm{\Theta}$ denotes the visual encoder parameters, $\mathcal T$ is a fixed input transformation, and $\mathcal H$ maps visual representations to class scores.
Let $\mathcal F_{\bm{\Theta}}$ be the function extracting global and intermediate visual representations.
The predicted class $\widehat y_{\mathcal M}(\bm{x})$ for image $\bm{x}$ is
\begin{equation}
\label{eq:ame-prediction}
\widehat y_{\mathcal M}(\bm{x})
=\argmax_{c\in\mathcal Y_T}
\big[\mathcal H(\mathcal F_{\bm{\Theta}}(\mathcal T(\bm{x})))\big]_c.
\end{equation}
The classifier can incorporate class-name text embeddings from the VLM.
We measure target-domain performance by the classification risk
\begin{equation}
\label{eq:ame-target-risk}
\mathcal R_T(\mathcal M)
=\mathbb E_{(\bm{x},y)\sim p_T}\big[\mathcal L(\widehat y_{\mathcal M}(\bm{x}),y)\big],
\end{equation}
where $\mathcal L(\widehat y,y)=\mathbf 1\{\widehat y\ne y\}$ is the classification error.
An executable adaptation algorithm $\mathcal A$ produces the adapted model, denoted as $\mathcal M_{\mathcal A}(\mathcal S)=\mathcal A(\mathcal M_0,\mathcal S,\mathcal C_T)$.

Conventional methods \cite{zhang2022TipAdapter,wang2024GDA,ye2026CDPS} apply the same predefined adaptation algorithm $\mathcal A_{\mathrm{shared}}$ to every target domain:
\begin{equation}
\label{eq:ame-conventional-objective}
\mathcal M_T^{\mathrm{conv}}
=\mathcal A_{\mathrm{shared}}(\mathcal M_0,\mathcal S_T,\mathcal C_T).
\end{equation}

The fitted model can differ across domains, while $\mathcal A_{\mathrm{shared}}$ remains the same. \method instead generates and revises a target-conditioned program $\mathcal A_T$, with the objective
\begin{equation}
\label{eq:ame-objective}
\min_{\mathcal A_T\in\mathcal P}\;
\mathcal R_T\big(\mathcal M_{\mathcal A_T}(\mathcal S_T)\big),
\end{equation}
where $\mathcal P$ comprises executable procedures for fitting $(\bm{\Theta},\mathcal T,\mathcal H)$ from the support set.

\section{Method}
\label{sec:method}

\textbf{\textit{Framework Overview.}}
Figure~\ref{fig:framework} illustrates the overall framework of \method.
The pretrained VLM, labeled support examples, and class names are provided alongside tools for code execution, model access, and visualization (a).
The agent uses \emph{compute} and \emph{view} to inspect the images, class semantics, model parameters, intermediate representations, and activations (Sec.~\ref{sec:method-observations}).
It forms hypotheses from these observations, implements them through \emph{edit}, and evaluates the resulting programs on the support set (b; Sec.~\ref{sec:method-loops}).
The history of observations, programs, and experimental outcomes guides subsequent editing decisions.
Promising candidates are submitted to an external evaluator, whose aggregate validation accuracy guides further refinement and final selection (c; Sec.~\ref{sec:method-feedback}).
The selected model is frozen for deployment. The agent, instructions, and actions are shared across domains; each target domain develops its own research history and executable adaptation program.

\subsection{Agent Observations and Actions}
\label{sec:method-observations}
The agent observes both the labeled examples in $\mathcal S_T$ and the VLM's computation on these examples.
Specifically, it can inspect: \ding{172} the input images, their labels, and the class names, which reveal visual characteristics and category semantics; \ding{173} the names, shapes, and numerical properties of parameter tensors, which identify the components available for editing; \ding{174} activations and representations from different layers, including global features, patch tokens, and attention maps; and \ding{175} prediction scores and misclassified examples in the support set, which reveal the distinctions that the current model fails to capture.
The agent selects specific images, model components, and outputs to inspect according to the hypothesis under investigation.
These observations serve different roles: examples specify the recognition problem, whereas internal quantities identify candidate interventions in the model. The agent can relate an observed confusion to the computation that produces it.

To investigate and adapt the model, the agent can take four actions: \ding{172} \emph{compute} model predictions, intermediate representations, and parameter statistics, identifying misclassified examples; \ding{173} \emph{view} text, images, image crops, and visualizations of model parameters and activations; \ding{174} \emph{edit} the model, including input transformations $\mathcal T$, encoder weights $\bm \Theta$ and the classifier $\mathcal H$; \ding{175} \emph{submit} a candidate model for external validation to receive the aggregate accuracy or for final deployment.
To support these actions, we provide tools for code execution, model access, and visualization.
The agent writes a fitting procedure in which encoder edits, representation choices, and classifier construction can be revised together. Fitting uses forward computation and numerical methods; BP-based runs also permit gradients.

\subsection{Agent Experiment Loops}
\label{sec:method-loops}
\method alternates observation, hypothesis, editing, and evaluation to develop a complete adaptation program. Starting from $\mathcal M_0$, the agent inspects support examples and model computations through \emph{compute} and \emph{view}, implements a hypothesis through \emph{edit}, and tests the resulting program through \emph{compute}. Encoder changes alter the features available to the classifier; classification errors, in turn, guide further observation and editing. Each cycle can therefore revise the representation and decision rule together.

Let $\mathcal M_t$ denote the current model, $\mathcal B_t$ the research history, and $\mathcal I_T$ the task instructions.
At step $t$, the agent $\pi_{\bm{\Psi}}$ chooses an action $a_t$ from its history, executes it, and records the output:
\begin{equation}
\label{eq:ame-interaction}
\begin{aligned}
a_t&\sim\pi_{\bm{\Psi}}(\cdot\mid\mathcal I_T,\mathcal B_t),\\
o_{t+1}&=\mathcal G_T(a_t;\mathcal M_t,\mathcal B_t),\\
\mathcal B_{t+1}&=\mathcal B_t\oplus(a_t,o_{t+1}),
\end{aligned}
\end{equation}
where $\mathcal G_T$ executes the selected action using the provided tools, $o_{t+1}$ is its output, and $\oplus$ appends the action and output to the history.
The history retains hypotheses, executed programs, observations, and evaluation outcomes, including unsuccessful trials, so that each new experiment can build upon earlier findings.
An \emph{edit} action executes an agent-written program $\mathcal U_t$ to update the current model:
\begin{equation}
\label{eq:ame-model-transition}
\mathcal M_{t+1}=
\begin{cases}
\mathcal U_t(\mathcal M_t;\mathcal S_T,\mathcal C_T), & \text{if }a_t\text{ edits the model},\\
\mathcal M_t, & \text{otherwise}.
\end{cases}
\end{equation}
The program may change any subset of $\bm{\Theta}$, $\mathcal T$, and $\mathcal H$, while the agent parameters $\bm{\Psi}$ remain fixed.
Subsequent \emph{compute} and \emph{view} actions examine the updated model, providing evidence for the next editing decision.

Evaluating an edit on the same images used to fit it can conceal overfitting.
To assess whether it also helps on examples excluded from fitting, the agent can use \emph{compute} to compare adaptation programs through cross-validation within the support set.
Specifically, partition $\mathcal S_T$ into disjoint folds $\{\mathcal S_T^{(b)}\}_{b=1}^{B}$ and write $\mathcal S_{T,-b}=\mathcal S_T\setminus\mathcal S_T^{(b)}$.
For each fold, the complete program $\mathcal A$ is fitted using only $\mathcal S_{T,-b}$, including weight edits, feature statistics, and classifier fitting, then evaluated on $\mathcal S_T^{(b)}$.
Aggregating the held-out predictions gives the estimated classification error
\begin{equation}
\label{eq:ame-support-evaluation}
\begin{aligned}
\widehat{\mathcal R}_{\mathrm{sup}}(\mathcal A)
=\frac{1}{n_T}\sum_{b=1}^{B}
\sum_{(\bm{x},y)\in\mathcal S_T^{(b)}}
\mathcal L\big(\widehat y_{\mathcal M_{\mathcal A}(\mathcal S_{T,-b})}(\bm{x}),y\big).
\end{aligned}
\end{equation}
The agent combines this error with predictions and visualizations from the held-out examples to decide whether to retain an edit, revise its implementation, or revisit the underlying hypothesis.
It then returns to observation and editing to test the revised program, closing the experiment loop.
Because successive designs use the same support examples, these comparisons provide local diagnostic evidence rather than an independent performance estimate.
Once a program $\mathcal A_j$ is promising, the agent reruns its complete fitting procedure on the full support set to produce $\mathcal M_j=\mathcal M_{\mathcal A_j}(\mathcal S_T)$ and uses \emph{submit} for external validation.

\subsection{External Feedback}
\label{sec:method-feedback}
Support-set experiments provide detailed diagnosis. External feedback evaluates submitted candidates on separate examples and supplies a common criterion for program refinement and final selection.
A validation \emph{submit} action records a candidate $\mathcal M_j$ and returns its aggregate accuracy on a separate validation set $\mathcal V_T$:
\begin{equation}
\label{eq:ame-external-feedback}
s_j=\frac{1}{|\mathcal V_T|}
\sum_{(\bm{x},y)\in\mathcal V_T}
\mathbf 1\{\widehat y_{\mathcal M_j}(\bm{x})=y\}.
\end{equation}
The agent receives this score but cannot inspect validation images, labels, representations, or individual predictions.
The score $s_j$ enters the research history as the output of the \emph{submit} action.
Program fitting and detailed diagnosis continue to use only the support set.
After receiving $s_j$, the agent can use \emph{compute} and \emph{view} to investigate the candidate, then \emph{edit} to revise it through the interaction in Eq.~\ref{eq:ame-interaction}.

Denoting $\mathcal J_T$ as the set of indices of valid externally evaluated candidates,
the final model is selected as
\begin{equation}
\label{eq:ame-selection}
j^\star\in\argmax_{j\in\mathcal J_T}s_j,
\qquad
\mathcal M_T^\star=\mathcal M_{j^\star}.
\end{equation}
The final deployment submission selects this evaluated candidate and freezes its parameters, input transformation, and classifier.
Test images and test outcomes do not enter the agent's history, and no further adaptation is performed on individual test images.
The shared agentic framework thus produces a domain-customized adaptation program and one fixed predictor for subsequent images from that domain.

\subsection{Summary}
\label{sec:method-summary}
Algorithm~\ref{alg:overall} summarizes \method. The agent uses observations and support-set experiments to revise its adaptation program, while aggregate external feedback guides refinement and ranks submitted candidates. The best evaluated model is then frozen for subsequent inference.

\begin{algorithm}[t]\small
\DontPrintSemicolon
\SetAlgoLined
\SetCommentSty{algorithmcomment}
\LinesNumbered
\SetKw{KwBreak}{break}
\KwIn{Pretrained VLM $\mathcal M_0$, support set $\mathcal S_T$, class names $\mathcal C_T$, instructions $\mathcal I_T$, agent $\pi_{\bm{\Psi}}$, tools $\mathcal G_T$.}
\KwOut{The target-domain model $\mathcal M_T^\star$.}
Initialize history $\mathcal B_0\gets\varnothing$, candidate set $\mathcal J_T\gets\varnothing$, and $t\gets q\gets0$\;
\While{no deployment request and $q<5$}{
  $a_t\sim\pi_{\bm{\Psi}}(\cdot\mid\mathcal I_T,\mathcal B_t)$ \tcp*{Choose next action}
  Set $\mathcal M_{t+1}\gets\mathcal M_t$ \tcp*{Carry current model forward}
  \uIf(\tcp*[f]{Observe evidence}){$a_t$ is \emph{compute} or \emph{view}}{
    Obtain $o_{t+1}$ from $\mathcal M_t$ and $\mathcal S_T$ (Sec.~\ref{sec:method-observations})\;
  }
  \uElseIf(\tcp*[f]{Implement hypothesis}){$a_t$ is \emph{edit}}{
    Execute the program $\mathcal U_t$ using $\mathcal S_T$ and $\mathcal C_T$\;
    Obtain $\mathcal M_{t+1}$ by Eq.~\ref{eq:ame-model-transition} and execution output $o_{t+1}$\;
  }
  \uElseIf{$a_t$ submits for validation}{
    $q\gets q+1$; fit and save $\mathcal M_j\gets\mathcal M_{\mathcal A_j}(\mathcal S_T)$\;
    Set $o_{t+1}$ to validation output \tcp*{Feedback}
    \lIf{valid}{record $s_j$; $\mathcal J_T\gets\mathcal J_T\cup\{j\}$}
  }
  \lElseIf{$a_t$ requests final deployment}{\KwBreak}
  Update $\mathcal B_{t+1}\gets\mathcal B_t\oplus(a_t,o_{t+1})$; $t\gets t+1$
}
Select $j^\star\in\argmax_{j\in\mathcal J_T}s_j$ and set $\mathcal M_{j^\star}$ as $\mathcal M_T^\star$\;
\Return $\mathcal M_T^\star$ \tcp*{The adapted model for deployment}
\caption{The overall procedure of \method.}
\label{alg:overall}
\end{algorithm}

%% file: 4_experiments.tex
\section{Experiments}
\label{sec:experiments}
\input{tables/main_results_bp_free}
\input{tables/main_results_bp_allowed}
\subsection{Experimental Setup}
\label{sec:setup}
\noindent\textbf{\textit{Datasets.}}
We evaluate \method on eleven datasets: Caltech101~\cite{li2004Caltech101}, OxfordPets~\cite{parkhi2012OxfordPets}, StanfordCars~\cite{krause2013StanfordCars}, Flowers102~\cite{nilsback2008Flowers102}, Food101~\cite{bossard2014Food101}, SUN397~\cite{xiao2010SUN}, DTD~\cite{cimpoi2014DTD}, EuroSAT~\cite{helber2019EuroSAT}, UCF101~\cite{soomro2012UCF101}, FGVCAircraft~\cite{maji2013FGVCAircraft}, and ImageNet~\cite{deng2009ImageNet}.
These datasets cover general object recognition, fine-grained classification, food, scenes, textures, remote sensing, and human actions.
We follow the public dataset splits and few-shot support lists, with Caltech101 using the corresponding 100-class version~\cite{zhou2022CoOp}. %

\smallskip
\noindent\textbf{\textit{Compared Baselines.}}
For BP-free baselines, we adopt Tip-Adapter~\cite{zhang2022TipAdapter}, APE~\cite{zhu2023APE}, TIP-X~\cite{udandarao2023TIPX}, GDA~\cite{wang2024GDA}, LBBO~\cite{liu2024LBBO}, IPO~\cite{du2024IPO}, TIMO~\cite{li2025TIMO}, and CDPS~\cite{ye2026CDPS}.
For \modelname{Qwen3-VL}, we additionally include SAVs~\cite{mitra2025SAVs} and MPO~\cite{choi2026MPO} as BP-free baselines.
For BP-based baselines, we use Tip-Adapter-F~\cite{zhang2022TipAdapter}, CoOp~\cite{zhou2022CoOp}, MaPLe~\cite{khattak2023MaPLe}, CLAP~\cite{silvarodriguez2024CLAP}, 2SFS~\cite{farina20252SFS}, and ProLIP~\cite{fahes2026ProLIP}. 
We also include the zero-shot backbone as an unadapted reference.

\smallskip
\noindent\textbf{\textit{Implementation Details.}}
We use \modelname{CLIP ViT-B/16}~\cite{radford2021learning} as the default backbone and \modelname{GPT-6.1-sol}~\cite{openai2026GPT61Sol} with high reasoning effort as the LLM agent. LLM-based baselines use the same LLM backend.
Each run uses $k=8$ labeled support images per class and five cumulative validation submissions.
Only aggregate validation accuracy is returned and the selected model is frozen before testing.
CLIP uses one $224\times224$ view and one visual encoder forward pass per image.
All methods share the backbone, support sets, validation protocol, and test split.
For each dataset and method, we report mean accuracy and sample standard deviation across three independently sampled support sets.

\subsection{Main Results}
\label{sec:main-results}
Table~\ref{tab:main-bp-free} compares BP-free adaptation across eleven domains: \textbf{\textit{Obs.\ding{182} \method achieves the highest average accuracy of $81.2\%$, compared with $79.8\%$ for the strongest baseline, TIMO.}}
On EuroSAT, it achieves a $6.2\%$ relative improvement over the strongest baseline.
\method matches or exceeds the strongest baseline on every dataset in Table~\ref{tab:main-bp-free}, covering fine-grained objects, textures, satellite imagery, and human actions.
For LLM-based prompt search \cite{liu2024LBBO,du2024IPO}, LBBO and IPO obtain average accuracies of $66.7\%$ and $67.8\%$, compared with $81.2\%$ for \method under the same LLM backend.
These methods search textual prompts for a fixed VLM; \method allows input transformations, encoder edits, and classifier construction. The comparison supports the utility of this broader adaptation space: \textbf{\textit{Obs.\ding{183} \method achieves higher average accuracy than these prompt-search methods.}}

\subsection{Further Analysis}
\textbf{\textit{Performance with Back-Propagation Enabled.}}
We also evaluate \method with gradient-based model updates enabled.
As shown in Table~\ref{tab:main-bp-based}, \textbf{\textit{Obs.\ding{184} \method achieves the highest average accuracy of $81.9\%$, compared with $80.6\%$ for the strongest baseline, 2SFS.}}
On DTD, it achieves a $4.6\%$ relative improvement over the strongest baseline.
The framework supports both forward-only numerical fitting and gradient-based parameter updates.

\begin{figure}[t]
    \centering
    \includegraphics[width=\linewidth]{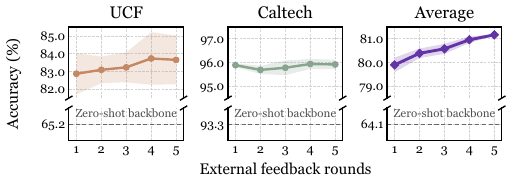}
    \vspace{-5mm}
    \caption{Effects of external feedback on UCF101, Caltech101, and the eleven-domain average. At each round, validation accuracy selects among candidates submitted so far; curves show their test accuracies with one standard deviation. The initial program is submitted before receiving external feedback.}
    \vspace{-3mm}
    \label{fig:external-feedback}
\end{figure}

\smallskip\noindent
\textbf{\textit{Effects of External Feedback.}}
We retrospectively truncate each recorded trajectory after one to five submissions and select the candidate with the highest validation accuracy within each prefix. The test set is used only to report the selected candidate's accuracy.
Round~1 therefore measures the first submitted program, before external feedback can influence subsequent design. Figure~\ref{fig:external-feedback} shows that this initial program already exceeds the strongest BP-free baseline in eleven-domain average accuracy.
Average accuracy rises over successive submissions, with different trajectories across domains: UCF improves over successive submissions, whereas Caltech remains close to its strong initial result.
Thus, \textbf{\textit{Obs.\ding{185} \method generates effective initial programs, and average accuracy increases along the recorded feedback trajectories, with gains varying across domains.}}

\input{tables/results_siglip2}
\input{tables/results_qwen3vl}

\smallskip\noindent
\textbf{\textit{Results on Different VLMs.}}
To test adaptation beyond the default backbone, we evaluate a contrastive VLM \modelname{SigLIP2-Base}~\cite{tschannen2025SigLIP2} and an instruction-tuned generative VLM \modelname{Qwen3-VL-4B-Instruct}~\cite{bai2025Qwen3VL}.
Both extensions use BP-free adaptation with $k=8$ support images per class over three support sets on DTD, Caltech, Aircraft, EuroSAT, and Cars.
As shown in Tables~\ref{tab:vlm-siglip2} and~\ref{tab:vlm-qwen3vl}, \method obtains the highest accuracy on all five datasets for each backbone, with average accuracies of $85.6\%$ and $85.2\%$, respectively.
On \modelname{SigLIP2}, the relative improvements over the strongest baseline on each dataset reach $5.5\%$ on Aircraft and $6.2\%$ on EuroSAT.
On \modelname{Qwen3-VL}, \method achieves $85.2\%$ average accuracy versus $77.1\%$ for the strongest compared baseline, with relative gains of $26.7\%$ on Aircraft and $16.7\%$ on EuroSAT.
These results show that \textbf{\textit{Obs.\ding{186} \method supports few-shot adaptation across the evaluated contrastive and instruction-tuned generative VLM backbones.}}

\input{tables/results_llm_agents}

\smallskip\noindent
\textbf{\textit{Results on Different LLM Agent Backends.}}
We replace the default LLM agent with \modelname{GPT-6.1-terra} and \modelname{GPT-6-luna}, keeping \modelname{CLIP ViT-B/16}, the instructions, tools, support sets, and BP-free adaptation protocol fixed.
Table~\ref{tab:llm-agents} reports results on DTD, Caltech, Aircraft, EuroSAT, and Cars.
All three agents improve accuracy on every dataset, with five-domain averages of $76.6\%$, $75.3\%$, and $74.8\%$ for \modelname{GPT-6.1-sol}, \modelname{GPT-6.1-terra}, and \modelname{GPT-6-luna}, respectively, compared with $53.8\%$ for zero-shot.
The results show that \textbf{\textit{Obs.\ding{187} \method remains effective across the evaluated LLM agents, while the choice of agent affects adaptation quality.}}
For example, results vary more on EuroSAT than on Caltech, and the default \modelname{GPT-6.1-sol} obtains the highest average accuracy.

\begin{figure}[t]
    \centering
    \includegraphics[width=\linewidth]{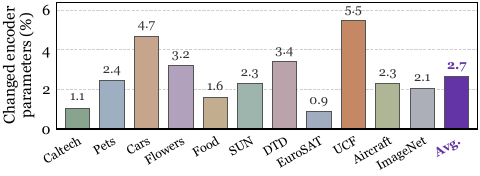}
    \vspace{-5mm}
    \caption{Percentage of visual-encoder parameters changed in the final adapted model. Most pretrained parameters remain unchanged, and the editing ratio varies across domains.}
    \vspace{-2mm}
    \label{fig:encoder-editing-ratio}
\end{figure}

\smallskip\noindent
\textbf{\textit{Model Editing Statistics.}}
We quantify editing as the fraction of visual-encoder scalar parameters whose deployed values differ from the original \modelname{CLIP ViT-B/16} weights, including the final projection.
Figure~\ref{fig:encoder-editing-ratio} summarizes the final BP-free models across three support sets per domain, counting unmodified encoders as zero.
These results show that \textbf{\textit{Obs.\ding{188} the final models retain most pretrained parameters, while the extent of editing varies across domains.}}
The mean editing ratio is $2.7\%$, with domain means ranging from $0.9\%$ on EuroSAT to $5.5\%$ on UCF.

\setlength{\textfloatsep}{20pt}
\begin{figure}[!t]
\centering
\includegraphics[width=\linewidth]{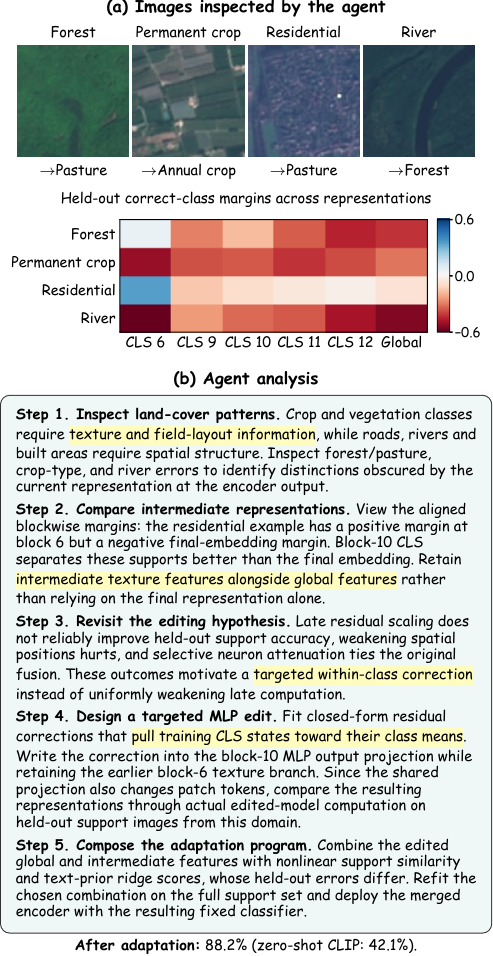}
\caption{Single-run EuroSAT case with \modelname{CLIP}. (a) Misclassified support images and margins across encoder layers; blue/red indicate correct/incorrect predictions. (b) Summary connecting layer representations, an MLP correction, and feature fusion.}
\vspace{-4mm}
\label{fig:case-eurosat}
\end{figure}

\begin{figure}[!t]
\centering
\includegraphics[width=\linewidth]{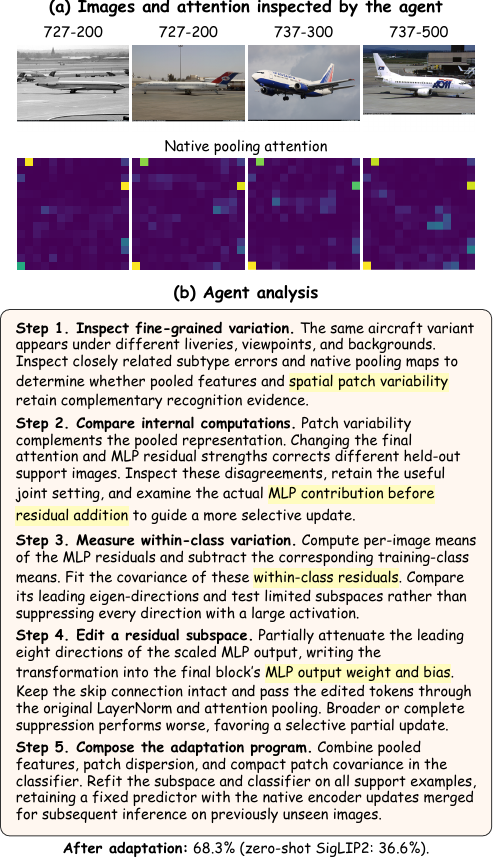}
\caption{Single-run Aircraft case with \modelname{SigLIP2}. (a) Inspected support images and native pooling attention. (b) Summarized agent analysis connecting within-class MLP residual variation with a rank-8 encoder edit and feature fusion.}
\vspace{-4mm}
\label{fig:case-aircraft}
\end{figure}

\subsection{Case Studies}
\label{sec:case-study}
Figures~\ref{fig:case-eurosat} and~\ref{fig:case-aircraft} summarize how observations lead to executable edits in two adaptation runs: EuroSAT (\modelname{CLIP}) and Aircraft (\modelname{SigLIP2}).
On EuroSAT, the agent inspects land-cover errors across encoder layers.
Block-6 features correctly classify a residential example that the final embedding misclassifies, revealing useful evidence in an intermediate layer.
After residual scaling and neuron attenuation fail to improve the support results, the agent fits a block-10 MLP correction with target $\Delta\bm z_i=0.25(\bm\mu_{y_i}-\bm z_i)$, where $\bm z_i$ is a CLS state and $\bm\mu_{y_i}$ its class mean.
The MLP correction targets within-class variation in later states, while the block-6 branch retains earlier texture evidence. Since the projection is shared by CLS and patch tokens, the agent evaluates the complete edited encoder. Its global and intermediate features enter nonlinear support kernels and text-prior ridge classifiers.
The complete program achieves $88.2\%$ test accuracy versus $42.1\%$ for zero-shot \modelname{CLIP}.%

On Aircraft, the agent inspects subtype confusions under changing liveries, viewpoints, and backgrounds.
Pooling maps and patch statistics suggest complementary spatial evidence; comparisons of residual edits then guide inspection of the MLP contribution.
It then measures within-class variation in the final block's MLP contributions and fits a rank-8 partial contraction.
Subtracting each subtype's training mean isolates within-class variation for the subspace fit. The update is written into the native MLP weight and bias, preserving the skip connection while changing tokens before LayerNorm and attention pooling. The classifier combines pooled features with patch dispersion and compact patch covariance from the edited forward pass.
The complete adapted model achieves $68.3\%$ test accuracy, compared with $36.6\%$ for zero-shot \modelname{SigLIP2}.
EuroSAT combines block-6 texture features with the global and block-10 representations affected by the correction, which propagates through the remaining Transformer blocks. Aircraft combines pooled and spatial statistics after its residual-subspace edit passes through normalization and pooling. Both classifiers are fitted to the edited representations. These cases illustrate \textbf{\textit{Obs.\ding{189} how \method connects domain-specific observations with executable programs for joint encoder adaptation and supervised classifier fitting.}}

%% file: tables/main_results_bp_free.tex
\begin{table*}[t]
\centering
\caption{BP-free few-shot adaptation accuracy (\%) using \modelname{CLIP ViT-B/16} with $k=8$ labeled images per class. \textbf{Bold} and \underline{underlined} values indicate the best and second-best displayed results, respectively.}
\vspace{-2mm}
\label{tab:main-bp-free}
\label{tab:summary} %
\begingroup
\fontsize{8}{9.2}\selectfont
\setlength{\tabcolsep}{3pt}
\renewcommand{\arraystretch}{1.12}
\resizebox{\textwidth}{!}{\begin{tabular}{l *{12}{c}}
\Xhline{1.2pt}
\rowcolor{ameTableHead!50}
Method & Caltech & Pets & Cars & Flowers & Food & SUN & DTD & EuroSAT & UCF & Aircraft & ImageNet & Avg. \\
\Xhline{1.0pt}
\rowcolor{ameTableStripe}
Zero-shot & 93.3 & 88.2 & 65.5 & 67.4 & 85.3 & 62.5 & 44.4 & 42.1 & 65.2 & 23.9 & 66.7 & 64.1 \\
\Xhline{0.6pt}
Tip-Adapter & \ameScore{94.3}{0.1} & \ameScore{90.7}{0.8} & \ameScore{72.0}{0.5} & \ameScore{93.5}{0.8} & \ameScore{86.1}{0.3} & \ameScore{70.5}{0.3} & \ameScore{62.5}{1.2} & \ameScore{71.8}{2.5} & \ameScore{76.0}{0.4} & \ameScore{35.6}{1.3} & \ameScore{70.1}{0.1} & \ameScore{74.8}{0.2} \\
\rowcolor{ameTableStripe}
APE & \ameScore{95.3}{0.1} & \ameScore{91.6}{0.5} & \ameScore{75.9}{0.6} & \ameScore{94.6}{0.4} & \ameScore{86.5}{0.1} & \ameScore{73.6}{0.2} & \ameScore{66.8}{3.0} & \ameScore{73.4}{2.9} & \ameScore{80.1}{1.5} & \ameScore{38.5}{0.5} & \ameScore{71.4}{0.1} & \ameScore{77.0}{0.5} \\
TIP-X & \ameScore{94.7}{0.6} & \ameScore{91.9}{1.0} & \ameScore{75.0}{0.3} & \ameScore{93.9}{0.7} & \ameScore{86.3}{0.1} & \ameScore{71.7}{0.4} & \ameScore{63.8}{0.1} & \ameScore{73.2}{3.6} & \ameScore{78.0}{0.4} & \ameScore{37.9}{0.7} & \ameScore{70.2}{0.2} & \ameScore{76.1}{0.2} \\
\rowcolor{ameTableStripe}
GDA & \ameScore{94.6}{0.5} & \ameScore{\underline{92.1}}{0.1} & \ameScore{77.8}{0.2} & \ameScore{97.2}{0.0} & \ameScore{86.7}{0.3} & \ameScore{\underline{74.0}}{0.2} & \ameScore{67.3}{0.7} & \ameScore{83.5}{3.5} & \ameScore{\underline{82.2}}{1.0} & \ameScore{43.8}{1.1} & \ameScore{71.5}{0.1} & \ameScore{79.2}{0.3} \\
LBBO & \ameScore{93.6}{0.0} & \ameScore{91.3}{0.2} & \ameScore{66.5}{0.8} & \ameScore{72.1}{0.4} & \ameScore{86.1}{0.1} & \ameScore{65.1}{0.1} & \ameScore{46.7}{0.8} & \ameScore{50.8}{2.1} & \ameScore{67.7}{0.1} & \ameScore{25.2}{0.7} & \ameScore{68.1}{0.3} & \ameScore{66.7}{0.3} \\
\rowcolor{ameTableStripe}
IPO & \ameScore{94.7}{0.6} & \ameScore{91.2}{0.5} & \ameScore{66.2}{0.3} & \ameScore{72.5}{0.2} & \ameScore{86.6}{0.1} & \ameScore{67.2}{0.2} & \ameScore{46.4}{0.5} & \ameScore{58.6}{1.7} & \ameScore{69.0}{0.5} & \ameScore{25.1}{0.5} & \ameScore{68.6}{0.0} & \ameScore{67.8}{0.1} \\
TIMO & \ameScore{\underline{95.7}}{0.6} & \ameScore{\underline{92.1}}{2.1} & \ameScore{\underline{78.6}}{0.3} & \ameScore{\underline{97.7}}{0.1} & \ameScore{\underline{87.2}}{0.2} & \ameScore{\textbf{75.4}}{0.3} & \ameScore{\underline{69.2}}{0.6} & \ameScore{\underline{84.3}}{2.6} & \ameScore{81.7}{1.1} & \ameScore{\underline{44.1}}{0.7} & \ameScore{\underline{72.3}}{0.1} & \ameScore{\underline{79.8}}{0.2} \\
\rowcolor{ameTableStripe}
CDPS & \ameScore{94.9}{0.3} & \ameScore{90.7}{0.6} & \ameScore{76.4}{1.0} & \ameScore{94.7}{0.4} & \ameScore{84.4}{0.1} & \ameScore{71.2}{0.0} & \ameScore{65.8}{1.1} & \ameScore{79.1}{0.7} & \ameScore{79.2}{1.1} & \ameScore{42.0}{0.9} & \ameScore{69.7}{0.1} & \ameScore{77.1}{0.1} \\
\Xhline{1.0pt}
\rowcolor{ameTableOurs!65}
\textbf{Ours} & \ameScore{\textbf{95.9}}{0.2} & \ameScore{\textbf{93.2}}{0.8} & \ameScore{\textbf{80.5}}{0.2} & \ameScore{\textbf{97.8}}{0.2} & \ameScore{\textbf{87.3}}{0.2} & \ameScore{\textbf{75.4}}{0.5} & \ameScore{\textbf{71.3}}{0.7} & \ameScore{\textbf{89.5}}{1.4} & \ameScore{\textbf{83.7}}{1.4} & \ameScore{\textbf{46.0}}{0.9} & \ameScore{\textbf{72.4}}{0.2} & \ameScore{\textbf{81.2}}{0.0} \\
\Xhline{1.2pt}
\end{tabular}}
\endgroup
\vspace{-2mm}
\end{table*}

%% file: tables/main_results_bp_allowed.tex
\begin{table*}[t]
\centering
\caption{Few-shot adaptation accuracy (\%) with back-propagation enabled using \modelname{CLIP ViT-B/16} with $k=8$ labeled images per class. \textbf{Bold} and \underline{underlined} values indicate the best and second-best displayed results, respectively.}
\vspace{-2mm}
\label{tab:main-bp-based}
\label{tab:main-bp-allowed} %
\begingroup
\fontsize{8}{9.2}\selectfont
\setlength{\tabcolsep}{3pt}
\renewcommand{\arraystretch}{1.12}
\resizebox{\textwidth}{!}{\begin{tabular}{l *{12}{c}}
\Xhline{1.2pt}
\rowcolor{ameTableHead!50}
Method & Caltech & Pets & Cars & Flowers & Food & SUN & DTD & EuroSAT & UCF & Aircraft & ImageNet & Avg. \\
\Xhline{1.0pt}
\rowcolor{ameTableStripe}
Zero-shot & 93.3 & 88.2 & 65.5 & 67.4 & 85.3 & 62.5 & 44.4 & 42.1 & 65.2 & 23.9 & 66.7 & 64.1 \\
\Xhline{0.6pt}
Tip-Adapter-F & \ameScore{95.0}{0.2} & \ameScore{91.8}{0.4} & \ameScore{78.1}{0.9} & \ameScore{94.6}{0.5} & \ameScore{86.7}{0.2} & \ameScore{73.2}{0.4} & \ameScore{66.9}{0.8} & \ameScore{77.6}{4.5} & \ameScore{82.1}{1.1} & \ameScore{38.8}{1.4} & \ameScore{71.6}{0.1} & \ameScore{77.9}{0.3} \\
\rowcolor{ameTableStripe}
CoOp & \ameScore{94.3}{0.4} & \ameScore{91.5}{0.9} & \ameScore{79.4}{0.7} & \ameScore{94.7}{0.4} & \ameScore{82.9}{0.4} & \ameScore{71.7}{0.6} & \ameScore{65.1}{1.2} & \ameScore{77.7}{3.2} & \ameScore{80.1}{0.8} & \ameScore{38.6}{1.1} & \ameScore{69.6}{0.1} & \ameScore{76.9}{0.1} \\
MaPLe & \ameScore{95.1}{0.4} & \ameScore{\textbf{93.1}}{0.5} & \ameScore{71.3}{0.4} & \ameScore{89.2}{0.7} & \ameScore{\underline{87.3}}{0.1} & \ameScore{73.1}{0.3} & \ameScore{62.3}{0.7} & \ameScore{78.5}{1.8} & \ameScore{79.2}{0.5} & \ameScore{33.2}{0.6} & \ameScore{71.4}{0.2} & \ameScore{75.8}{0.1} \\
\rowcolor{ameTableStripe}
CLAP & \ameScore{95.2}{0.2} & \ameScore{92.5}{0.3} & \ameScore{79.0}{0.3} & \ameScore{95.4}{0.5} & \ameScore{87.0}{0.1} & \ameScore{74.5}{0.3} & \ameScore{67.8}{0.2} & \ameScore{77.1}{4.6} & \ameScore{82.7}{0.7} & \ameScore{40.7}{0.8} & \ameScore{71.4}{0.0} & \ameScore{78.5}{0.4} \\
2SFS & \ameScore{\underline{95.9}}{0.4} & \ameScore{\textbf{93.1}}{0.2} & \ameScore{\underline{81.3}}{0.6} & \ameScore{\underline{96.3}}{0.1} & \ameScore{86.3}{0.4} & \ameScore{\underline{75.6}}{0.2} & \ameScore{\underline{69.3}}{0.7} & \ameScore{\underline{88.0}}{2.4} & \ameScore{\underline{84.4}}{1.1} & \ameScore{\underline{44.6}}{1.0} & \ameScore{\underline{72.3}}{0.0} & \ameScore{\underline{80.6}}{0.1} \\
\rowcolor{ameTableStripe}
ProLIP & \ameScore{95.5}{0.2} & \ameScore{92.5}{0.3} & \ameScore{77.6}{0.1} & \ameScore{94.2}{0.5} & \ameScore{86.5}{0.2} & \ameScore{73.9}{0.2} & \ameScore{67.7}{0.2} & \ameScore{83.8}{3.0} & \ameScore{83.0}{0.7} & \ameScore{41.0}{0.6} & \ameScore{70.8}{0.1} & \ameScore{78.8}{0.3} \\
\Xhline{1.0pt}
\rowcolor{ameTableOurs!65}
\textbf{Ours} & \ameScore{\textbf{96.0}}{0.3} & \ameScore{\underline{92.9}}{0.8} & \ameScore{\textbf{82.2}}{1.2} & \ameScore{\textbf{97.8}}{0.1} & \ameScore{\textbf{87.4}}{0.2} & \ameScore{\textbf{76.6}}{0.1} & \ameScore{\textbf{72.5}}{0.8} & \ameScore{\textbf{90.4}}{1.9} & \ameScore{\textbf{85.4}}{1.2} & \ameScore{\textbf{47.5}}{0.7} & \ameScore{\textbf{72.6}}{0.4} & \ameScore{\textbf{81.9}}{0.1} \\
\Xhline{1.2pt}
\end{tabular}}
\endgroup
\vspace{-2mm}
\end{table*}

%% file: tables/results_siglip2.tex
\begin{table}[t]
\centering
\captionsetup{justification=raggedright,singlelinecheck=false}
\caption{Adaptation results of \method on another contrastive VLM, \modelname{SigLIP2-Base/16}, under the BP-free setting averaged across three random seeds. We \textbf{bold} the best and \underline{underline} second-best.}
\vspace{-2mm}
\label{tab:vlm-siglip2}
\begingroup
\fontsize{8}{9.2}\selectfont
\setlength{\tabcolsep}{2pt}
\renewcommand{\arraystretch}{1.12}
\resizebox{\linewidth}{!}{\begin{tabular}{l *{6}{c}}
\Xhline{1.2pt}
\rowcolor{ameTableHead!50}
Method & DTD & Caltech & Aircraft & EuroSAT & Cars & Avg. \\
\Xhline{1.0pt}
\rowcolor{ameTableStripe}
Zero-shot & 61.4 & 97.5 & 36.6 & 39.0 & 85.0 & 63.9 \\
\Xhline{0.6pt}
APE & \ameScore{73.4}{1.6} & \ameScore{\underline{98.3}}{0.2} & \ameScore{59.0}{1.0} & \ameScore{73.6}{2.6} & \ameScore{90.2}{0.7} & \ameScore{78.9}{0.2} \\
\rowcolor{ameTableStripe}
GDA & \ameScore{74.7}{1.5} & \ameScore{98.2}{0.2} & \ameScore{\underline{66.0}}{0.8} & \ameScore{\underline{83.9}}{3.3} & \ameScore{\underline{92.8}}{0.6} & \ameScore{\underline{83.1}}{0.6} \\
TIMO & \ameScore{\underline{76.3}}{2.6} & \ameScore{\underline{98.3}}{0.2} & \ameScore{64.7}{0.8} & \ameScore{83.6}{2.9} & \ameScore{92.2}{0.1} & \ameScore{83.0}{0.1} \\
\rowcolor{ameTableStripe}
CDPS & \ameScore{75.0}{0.8} & \ameScore{98.1}{0.4} & \ameScore{61.8}{1.3} & \ameScore{74.4}{1.6} & \ameScore{88.4}{0.2} & \ameScore{79.5}{0.3} \\
\Xhline{1.0pt}
\rowcolor{ameTableOurs!65}
\textbf{Ours} & \ameScore{\textbf{78.0}}{0.8} & \ameScore{\textbf{98.4}}{0.2} & \ameScore{\textbf{69.6}}{2.0} & \ameScore{\textbf{89.1}}{1.4} & \ameScore{\textbf{92.9}}{0.7} & \ameScore{\textbf{85.6}}{0.7} \\
\Xhline{1.2pt}
\end{tabular}}
\endgroup
\vspace{-2mm}
\end{table}

%% file: tables/results_qwen3vl.tex
\begin{table}[t]
\centering
\captionsetup{justification=raggedright,singlelinecheck=false}
\caption{Adaptation results of \method on an instruction-following generative VLM, \modelname{Qwen3-VL-4B-Instruct}, under the BP-free setting averaged across three random seeds. The best and second-best results are \textbf{bold} and \underline{underlined}, respectively.}
\vspace{-2mm}
\label{tab:vlm-qwen3vl}
\begingroup
\fontsize{8}{9.2}\selectfont
\setlength{\tabcolsep}{2pt}
\renewcommand{\arraystretch}{1.12}
\resizebox{\linewidth}{!}{\begin{tabular}{l *{6}{c}}
\Xhline{1.2pt}
\rowcolor{ameTableHead!50}
Method & DTD & Caltech & Aircraft & EuroSAT & Cars & Avg. \\
\Xhline{1.0pt}
\rowcolor{ameTableStripe}
Zero-shot & 50.9 & 87.3 & 19.2 & 51.8 & 41.4 & 50.1 \\
\Xhline{0.6pt}
LBBO & \ameScore{50.7}{0.4} & \ameScore{87.2}{0.1} & \ameScore{20.3}{0.0} & \ameScore{52.3}{0.2} & \ameScore{41.0}{0.4} & \ameScore{50.3}{0.0} \\
\rowcolor{ameTableStripe}
IPO & \ameScore{50.6}{0.2} & \ameScore{87.5}{0.1} & \ameScore{20.0}{0.1} & \ameScore{54.1}{1.3} & \ameScore{42.4}{0.3} & \ameScore{50.9}{0.3} \\
SAVs & \ameScore{\underline{73.7}}{0.5} & \ameScore{\underline{97.2}}{0.1} & \ameScore{\underline{54.3}}{1.3} & \ameScore{\underline{76.8}}{3.3} & \ameScore{\underline{83.6}}{0.4} & \ameScore{\underline{77.1}}{0.6} \\
\rowcolor{ameTableStripe}
MPO & \ameScore{51.4}{0.9} & \ameScore{89.8}{2.2} & \ameScore{20.3}{0.5} & \ameScore{55.6}{2.0} & \ameScore{40.8}{0.9} & \ameScore{51.6}{0.6} \\
\Xhline{1.0pt}
\rowcolor{ameTableOurs!65}
\textbf{Ours} & \ameScore{\textbf{77.2}}{0.9} & \ameScore{\textbf{98.4}}{0.3} & \ameScore{\textbf{68.8}}{1.7} & \ameScore{\textbf{89.6}}{1.3} & \ameScore{\textbf{92.1}}{0.1} & \ameScore{\textbf{85.2}}{0.6} \\
\Xhline{1.2pt}
\end{tabular}}
\endgroup
\end{table}

%% file: tables/results_llm_agents.tex
\begin{table}[!t]
\centering
\captionsetup{justification=raggedright,singlelinecheck=false}
\caption{Adaptation results on \modelname{CLIP ViT-B/16} under the BP-free setting using different LLM agent backends. The best and second-best results are \textbf{bold} and \underline{underlined}, respectively.}
\vspace{-2mm}
\label{tab:llm-agents}
\begingroup
\fontsize{8}{9.2}\selectfont
\setlength{\tabcolsep}{2pt}
\renewcommand{\arraystretch}{1.12}
\resizebox{\linewidth}{!}{\begin{tabular}{l *{6}{c}}
\Xhline{1.2pt}
\rowcolor{ameTableHead!50}
Method / Agent & DTD & Caltech & Aircraft & EuroSAT & Cars & Avg. \\
\Xhline{1.0pt}
\rowcolor{ameTableStripe}
Zero-shot & 44.4 & 93.3 & 23.9 & 42.1 & 65.5 & 53.8 \\
\Xhline{0.6pt}
\modelname{GPT-6-luna} & \ameScore{70.3}{1.8} & \ameScore{\underline{95.4}}{0.0} & \ameScore{\underline{42.8}}{1.9} & \ameScore{86.5}{1.1} & \ameScore{\underline{79.1}}{0.6} & \ameScore{74.8}{0.2} \\
\rowcolor{ameTableStripe}
\modelname{GPT-6.1-terra} & \ameScore{\underline{70.8}}{2.3} & \ameScore{\textbf{95.9}}{0.2} & \ameScore{42.6}{3.1} & \ameScore{\underline{88.9}}{0.5} & \ameScore{78.2}{0.7} & \ameScore{\underline{75.3}}{1.0} \\
\rowcolor{ameTableOurs!65}
\modelname{GPT-6.1-sol} & \ameScore{\textbf{71.3}}{0.7} & \ameScore{\textbf{95.9}}{0.2} & \ameScore{\textbf{46.0}}{0.9} & \ameScore{\textbf{89.5}}{1.4} & \ameScore{\textbf{80.5}}{0.2} & \ameScore{\textbf{76.6}}{0.3} \\
\Xhline{1.2pt}
\end{tabular}}
\endgroup
\vspace{-2mm}
\end{table}

%% file: 5_conclusion.tex
\section{Conclusion}
\label{sec:conclusion}

We study few-shot VLM adaptation as the design of a complete, domain-customized fitting program. \method connects multimodal and internal model observations with executable changes to encoder computation, feature extraction, and classifier fitting. The agent revises these choices together through support-set experiments and aggregate external feedback, producing a fixed predictor for subsequent target images. Experiments across eleven datasets and contrastive and generative VLMs show its effectiveness with and without BP. Experiments with different LLM agents demonstrate applicability across backends, while the case studies illustrate how domain observations guide the joint design of encoder edits and classifiers.